\documentclass{article}
\usepackage{arxiv}

\usepackage[utf8]{inputenc} % allow utf-8 input
\usepackage[T1]{fontenc}    % use 8-bit T1 fonts
\usepackage{hyperref}       % hyperlinks
\usepackage{url}            % simple URL typesetting
\usepackage{booktabs}       % professional-quality tables
\usepackage{amsfonts}       % blackboard math symbols
\usepackage{nicefrac}       % compact symbols for 1/2, etc.
\usepackage{microtype}      % microtypography
\usepackage{graphicx}
\usepackage{natbib}
\usepackage{doi}
\usepackage{amsmath}

\usepackage{subcaption}
\usepackage{xcolor}
\usepackage{placeins}
\usepackage{longtable}
\usepackage{tabularx}
\usepackage{tikz}
\usetikzlibrary{shapes.geometric, arrows.meta, positioning}

\tikzstyle{block} = [rectangle, draw, minimum height=1cm, minimum width=2.5cm, text centered, text width=3.2cm]
\tikzstyle{arrow} = [thick, ->, >=stealth]
\tikzstyle{bubble} = [cloud, draw, align=center, text width=3.5cm, fill=gray!10]

\title{iOSPointMapper: RealTime Pedestrian and Accessibility Mapping with mobile AI}

\author{
Himanshu Naidu\thanks{Correspondence to Himanshu Naidu (hnaidu36@uw.edu) and Anat Caspi (caspian@cs.washington.edu).} \quad
Yuxiang Zhang \quad
Sachin Mehta \quad
Anat Caspi\footnotemark[1] \\
University of Washington \\
Seattle, Washington, USA
}

\begin{document}
\maketitle

\begin{abstract}
Accurate, up-to-date sidewalk data is essential for building accessible and inclusive pedestrian infrastructure, yet current approaches to data collection are often costly, fragmented, and difficult to scale. We introduce \textit{iOSPointMapper}, a mobile application that enables real-time, privacy-conscious sidewalk mapping on the ground, using recent-generation iPhones and iPads. The system leverages on-device semantic segmentation, LiDAR-based depth estimation, and fused GPS/IMU data to detect and localize sidewalk-relevant features such as traffic signs, traffic lights and poles. To ensure transparency and improve data quality, iOSPointMapper incorporates a user-guided annotation interface for validating system outputs before submission. Collected data is anonymized and transmitted to the Transportation Data Exchange Initiative (TDEI), where it integrates seamlessly with broader multimodal transportation datasets. Detailed evaluations of the system’s feature detection and spatial mapping performance reveal the application's potential for enhanced pedestrian mapping. Together, these capabilities offer a scalable and user-centered approach to closing critical data gaps in pedestrian infrastructure.
\end{abstract}


\setcounter{section}{0}
\section{Introduction}

Sidewalks form a critical component in the Public Right of Way, enabling pedestrians to move safely through urban and suburban environments \citep{ABOUSENNA2022106548}. Yet, unlike roadways, which benefit from substantial investments in maintenance and monitoring, sidewalks often suffer from neglect - both in infrastructure and data \citep{disparity_sidewalks_v_roads, Kohler_2023}. This neglect disproportionately affects individuals who rely on sidewalks as their primary means of travel, including people with disabilities, older adults, and those without access to personal vehicles. The lack of high-quality, up-to-date sidewalk data hinders the ability of local governments, transit agencies, and non-profit organizations to make informed decisions around maintenance, compliance, and advocacy.

Despite the growing importance and demand of accessible, multimodal transportation networks, there remains several gaps in scalable, standardized systems for assessing and monitoring sidewalk environments. A centralized and reliable mechanism to collect and manage this data would benefit a wide array of stakeholders. Government agencies could use such systems to better prioritize infrastructure investments, enforce compliance with accessibility regulations such as the Americans with Disabilities Act \citep{americans_1991}, facilitate reliable transportation, and improve emergency response. Private organizations, such as delivery services and sidewalk robotics companies, could improve operational efficiency by better understanding traversability in pedestrian environments. Non-profit Organizations and Advocacy groups could be empowered with a data-driven approach to push for equitable investments in pedestrian infrastructure. Most importantly, communities would benefit from safer, more walkable neighborhoods and the increased mobility that comes with them.

One of the initiatives in response to this pressing need, is TDEI \citep{Hallenbeck2021-06-28}, led by the Taskar Center for Accessible Technology and funded by the US Department of Transportation under the ITS4US Deployment Program \citep{usdot_its4us}. TDEI enables the integration and sharing of data on pedestrian pathways, transit services, and dynamic mobility networks through a common platform. It also supports machine learning-based tools, such as Prophet \citep{zhang2024reliable}, which uses aerial imagery and rasterized road network topologies to detect sidewalk features. However, aerial methods are limited in scope due to occlusions, variable resolution, and difficulty capturing fine-grained detail in complex urban environments. In many cases, a ground-based approach is necessary to provide high-fidelity, real-time assessment of the pedestrian infrastructure. 

To address this gap within the scope of TDEI, we present iOSPointMapper, a mobile application designed to perform ground-level mapping of sidewalk environments. iOSPointMapper runs on recent-generation iPhones and iPads equipped with high-resolution cameras and LiDAR sensors. The application leverages these alongside GPS and inertial measurement unit (IMU) data to detect and identify features relevant to sidewalk accessibility. The application uses real-time, on-device convolutional neural networks to perform semantic segmentation and feature detection, minimizing latency while preserving user privacy. Only anonymized geometric and semantic data - such as the location and width of a sidewalk - is sent to a central data repository, ensuring that no imary or Personally Identifiable Information (PII) is transmitted during the process. Recognizing the importance of human oversight in AI-assisted systems, iOSPointMapper includes a built-in annotation view that allows users to vet the system’s output before submission. This helps increase the reliability of collected data while fostering transparency and trust, and empowering users to leverage scalable mapping efforts without relinquishing control. Once validated, the resulting information is uploaded to a specified workspace within the TDEI platform, enabling seamless integration with existing systems and facilitating data-driven transportation planning.

With this work, we aim to establish a scalable, privacy-conscious, and user-centered approach for sidewalk data collection and assessment. As part of the broader TDEI framework, iOSPointMapper can help close critical data gaps in pedestrian infrastructure by enabling community-informed, ground-level assessments at scale.

The remainder of the paper is structured as follows. Section \ref{related_work} reviews prior work in pedestrian environment mapping. Section \ref{methodology} outlines the iOSPointMapper system, detailing how the sidewalk environment features are detected, localized and validated on-device before being transmitted to the centralized data storage. Section \ref{experiments} describes the experimental setup used to evaluate the system, while Section \ref{results} presents the results, including resource utilization and mapping performance. Section \ref{discussion} discusses the broader implications of this work for accessible sidewalk mobility and outlines directions for future development. We end with a conclusion in Section \ref{conclusion}.

\section{Related Work}\label{related_work}

Recent advances in mobile hardware and efficient computer vision model architectures have made it possible to use mobile computer vision and mapping in diverse environments, especially urban settings. While most prior work focused on automobile environments, new and emerging research has begun to address unique challenges posed by pedestrian environments. However, there hasn’t been a production-ready mobile-based solution for urban scene understanding in pedestrian environments in relation to accessibility and sidewalk infrastructure.

\subsection{Computer Vision for Feature Detection on Edge Devices}

Computer vision has been central to urban scene understanding across domains such as autonomous driving \citep{janai2020computer}, augmented reality \citep{minaee2022modern}, and robotics \citep{ruiz2018survey}. Tasks like object detection, depth estimation, and semantic segmentation underpin perception pipelines for navigation and infrastructure assessment. In these domains, high-throughput models are often deployed on high-end GPU-equipped servers or vehicle-mounted platforms, where bandwidth and compute are less constrained.

Semantic segmentation, in particular, has played an important role in autonomous vehicles and smart city applications \citep{electronics10040471}. Semantic segmentation models — including Convolutional architectures such as DeepLabv3 \citep{chen2017rethinking}, DDRNet \citep{hong2021deep}, HRNet \citep{DBLP:journals/corr/abs-1908-07919}, as well as Transformer-based models like SegFormer \citep{xie2021segformer} — have consistently achieved state-of-the-art performance on urban scene understanding benchmarks. These benchmarks include Cityscapes ~\citep{7780719}, Mapillary Vistas \citep{MVD2017} and KITTI \citep{doi:10.1177/0278364913491297}. To support real-time performance in mobile and embedded contexts, numerous lightweight model architectures have been developed \citep{9952938}. Models such as BiSeNetv2 \citep{DBLP:journals/corr/abs-2004-02147} and ESPNetv2 \citep{DBLP:journals/corr/abs-1811-11431} employ compact feature extraction layers and optimized decoding paths, making them well-suited for edge deployment in applications like pedestrian environment mapping.

However, in the context of urban scene understanding, since these approaches typically rely on vehicle-mounted cameras, they are largely unsuitable for feature detection in pedestrian-centric environments. In this paper, we refer \textit{feature detection} to the process of identifying salient visual elements that are critical for interpreting and navigating complex urban environments. Widely used training datasets \citep{7780719,MVD2017,doi:10.1177/0278364913491297} — prioritize vehicular viewpoints, often lacking the perspective and features relevant to pedestrian accessibility. To bridge this gap, pedestrian-specific segmentation tasks require both appropriate model selection and dataset adaptation. In our work, we employ BiSeNetv2 for efficient on-device semantic segmentation of sidewalk-relevant features, including sidewalks, poles, and traffic signs. The model is trained on a hybrid dataset that includes both curated street-level imagery from a large-scale dataset \citep{MVD2017}, and a custom annotated pedestrian-perspective dataset, ensuring alignment with the deployment setting. The resulting pipeline supports robust and privacy-preserving recognition on consumer-grade mobile devices.

\subsection{Mapping the Environment for Sidewalk Navigation}

In contrast to vehicular mapping systems, which benefit from standardized sensors, higher altitudes, and large-scale public datasets, pedestrian mapping remains fragmented in both methodology and scope. Several recent projects have sought to address aspects of sidewalk environment mapping—particularly for accessibility—but few offer integrated, real-time solutions suitable for mobile deployment.

A substantial portion of existing pedestrian-related mapping efforts rely on aerial imagery, either solely or in combination with other data. For instance, PathwayBench \citep{zhang2024pathwaybench} and Tile2Net \citep{HOSSEINI2023101950} focus on segmenting pedestrian paths using aerial imagery. \citet{doi:10.1177/2399808321995817} combines aerial imagery with street-view images either for completion-oriented post-processing or validation. Tools such as Prophet \citep{zhang2024reliable} use satellite imagery with rasterized road networks to extract sidewalk topology. While these approaches demonstrate that aerial imagery can support scalable mapping of pedestrian infrastructure, they are limited by challenges. For instance, dense vegetation, urban clutter, and buildings frequently occlude key sidewalk features. The resolution of aerial imagery is often too coarse to capture details like curb ramps, obstructions, or surface conditions. Moreover, the slower update frequency for aerial imagery makes it unable to account for the rapidly changing nature of urban environments. 

Recognizing the limitations of aerial methods, several projects have explored ground-based approaches for capturing sidewalk-level data. 
OASIS \citep{zhang2023oasis} uses stereo video from chest-mounted ZED cameras to detect and localize sidewalks, barriers, and pedestrian-relevant features, and generating a routable pedestrian graph out of the data. While effective in controlled deployments, OASIS depends on specialized stereo camera hardware, which limits its accessibility for large-scale, crowd-sourced mapping efforts using consumer devices. Other efforts (e.g., Sidewalk AI Scanner of \citet{sidewalkaiscanner}) attempt to leverage commodity smartphones, allowing users to collect sidewalk imagery using mobile phones, which is then uploaded to a server for processing. Although this approach offers a scalable solution, it introduces privacy risks due to server-side image transmission and analysis. Moreover, there is no user-engagement beyond data upload, which limits the user-vetting potential.

To our knowledge, few existing systems provide real-time sidewalk mapping using on-device computer vision and sensor fusion on consumer mobile devices. iOSPointMapper builds on this gap by integrating RGB imagery, depth maps (via LiDAR), GPS location, and camera pose to extract and geo-locate sidewalk-relevant features. By enabling user vetting of detected features, the application balances model-driven detection with human oversight, aligning with practices in accessibility research that emphasize trust, transparency, and validation

\section{Methodology} \label{methodology}

\begin{figure}[t!]
  \centering
  \includegraphics[width=1.0\textwidth]{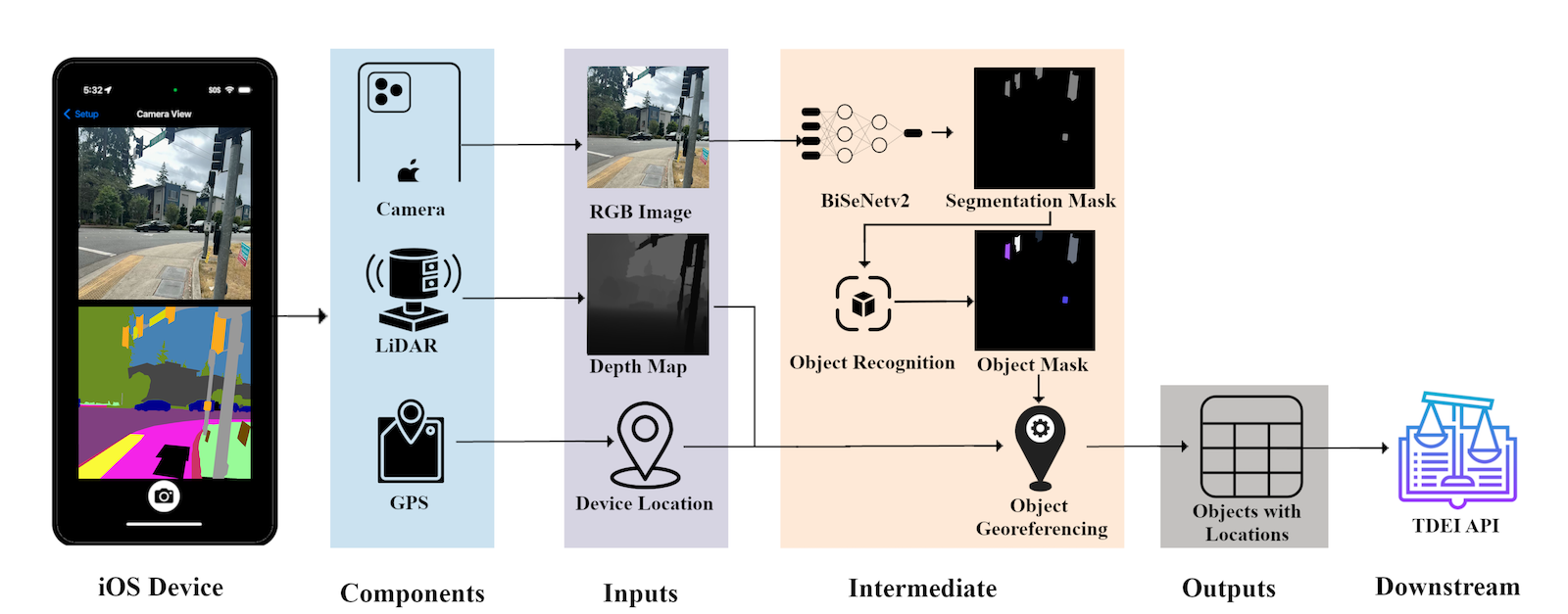}
  \caption{Schematic of iOSPointMapper}\label{fig:iOSPointMapper_schematic}
\end{figure}

This section describes the technical design and implementation of \textit{iOSPointMapper}, as shown in Figure \ref{fig:iOSPointMapper_schematic}. We begin by outlining the computer vision pipeline to detect and classify sidewalk features using images captured by the high-resolution camera in recent-generation iOS devices (Section \ref{sub_methodology_feature_recognition}). We then describe in Section \ref{sub_methodology_feature_localization} how these detected features are localized, using fused inputs from LiDAR, GPS and IMU sensors. Following this, we present the user-vetting interface that allows mappers to validate the captured features before submission (Section \ref{sub_methodology_vetting}). Finally, in Section \ref{sub_methodology_integration}, we detail how the vetted data is transmitted to a central TDEI workspace, ensuring immediate usability and standardized integration into broader transportation data systems. 

\subsection{Feature Detection in Sidewalk Environments} \label{sub_methodology_feature_recognition}

The feature detection pipeline at its core uses semantic segmentation, which enables real-time per-frame recognition of sidewalk-relevant object classes. The pipeline converts each camera frame into a dense segmentation mask, stabilizes predictions across frames, and post-processes the results into discrete features suitable for localization and mapping.

\subsubsection{Semantic Segmentation Modeling}
We use semantic segmentation to classify each pixel in the input RGB image to one of the categories (e.g., sidewalk and pole). Specifically, for a real-time semantic segmentation on iOS devices, 
%Semantic segmentation is a computer vision technique that assigns each pixel in an RGB image ($I \in \mathbb{Z}^{w \times h \times 3}$) an integer label corresponding to an object class. For a semantic class set $C$ that is assigned to the model during training, the model outputs a segmentation mask $S \in \mathbb{Z}^{w \times h}$, where each pixel $S_{ij}$ is obtained as:
%\[
%    S_{ij}=arg\_max_{c\in C}P(c\mid I_{ij})
%\]
%\hspace*{\parindent}To enable efficient real-time semantic segmentation on iOS devices, 
we use BiSeNetv2 \citep{DBLP:journals/corr/abs-2004-02147}, a lightweight convolutional neural network designed for real-time segmentation. The design    adopts a dual-branch architecture: a \emph{detail branch} for high-resolution spatial features and a \emph{semantic branch} for contextual understanding. This architecture is further enhanced by a booster training strategy, which improves segmentation quality during training without affecting inference time. BiSeNetv2 also works seamlessly with Apple’s CoreML framework, facilitating smooth conversion to a format suitable for iOS hardware. Its lightweight design and performance-oriented training strategy make it well-suited for mobile deployment, while maintaining competitive segmentation accuracy.

\subsubsection{Training Strategy}

The semantic segmentation model is trained to segment five sidewalk-relevant feature classes: \textit{sidewalk, building, traffic sign, traffic light, and pole}. These classes appear with  high frequency and visually variability across environments. They are seminally important for understanding sidewalk infrastructure and are often underrepresented in vehicular-centric datasets. In addition, background classes such as vegetation, terrain, and sky are segmented to reduce false positives and improve model robustness in complex environments.

Many existing semantic segmentation datasets—such as Cityscapes \citep{7780719}, KITTI \citep{doi:10.1177/0278364913491297}, and Mapillary Vistas \citep{MVD2017} — were primarily developed for automotive-centric applications. This focus limits their applicability for pedestrian-level semantic segmentation tasks, especially those requiring fine-grained recognition of sidewalk infrastructure and accessibility features. 

To address the limitations of these datasets, we curated a pedestrian-centric training corpus using a two-stage approach. First, we extracted a subset of the Mapillary Vistas dataset, prioritizing images with strong sidewalk presence, which resulted in a pre-training dataset of 4300 images. Second, we collected RGB-D video data across Seattle and Bellevue using LiDAR-enabled iPhones, mounted on a chest strap to emulate a pedestrian perspective. Still frames were extracted from the video at fixed intervals, and manually annotated to highlight sidewalk features under varied environments. The custom annotated images were then split into a training set of 200 images and a validation set of 50 images. 

In the training pipeline, we first pretrained BiSeNetv2 on the curated Mapillary Vistas subset, followed by fine-tuning on the custom annotated dataset. A selective fine-tuning strategy was adopted: the detail branch in BiSeNetv2 was frozen to preserve low-level feature extraction, while the semantic branch was trained to adapt the latent representations to our use case. This approach improved the model’s ability to localize small but critical features, such as poles and signage, while maintaining spatial accuracy across varied sidewalk geometries.

\subsubsection{Discrete Feature Extraction with Temporal Stabilization}

To enhance the segmentation stability, \textit{iOSPointMapper} leverages the temporal continuity of real-time video data. For each frame that the user chooses to capture, several preceding consecutive frames are also processed. These previous frames are aligned with the captured frame using homography transformation, and then the segmentation masks of these frames are merged via a per-pixel majority-vote scheme. Following this aggregation step, contour detection is applied to the final segmentation mask to extract discrete object instances for each class. This allows the system to convert dense pixel-wise predictions into discrete features suitable for localization and mapping.

\subsubsection{User Interface}

\begin{figure}[b!]
  \centering
    \begin{subfigure}[b]{0.3\columnwidth}
    \includegraphics[width=\columnwidth]{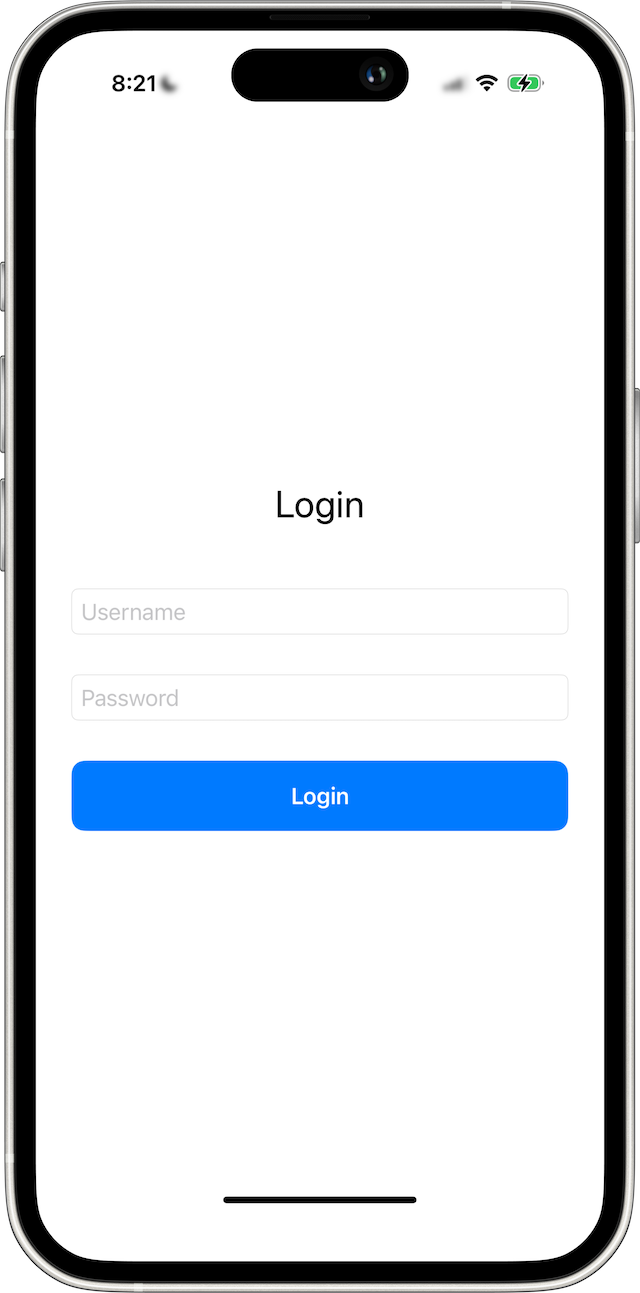}
    \caption{}
    \label{fig:login_setup_camera_screenshots_a}
    \end{subfigure}
    \hfill
    \begin{subfigure}[b]{0.3\columnwidth}
    \includegraphics[width=\columnwidth]{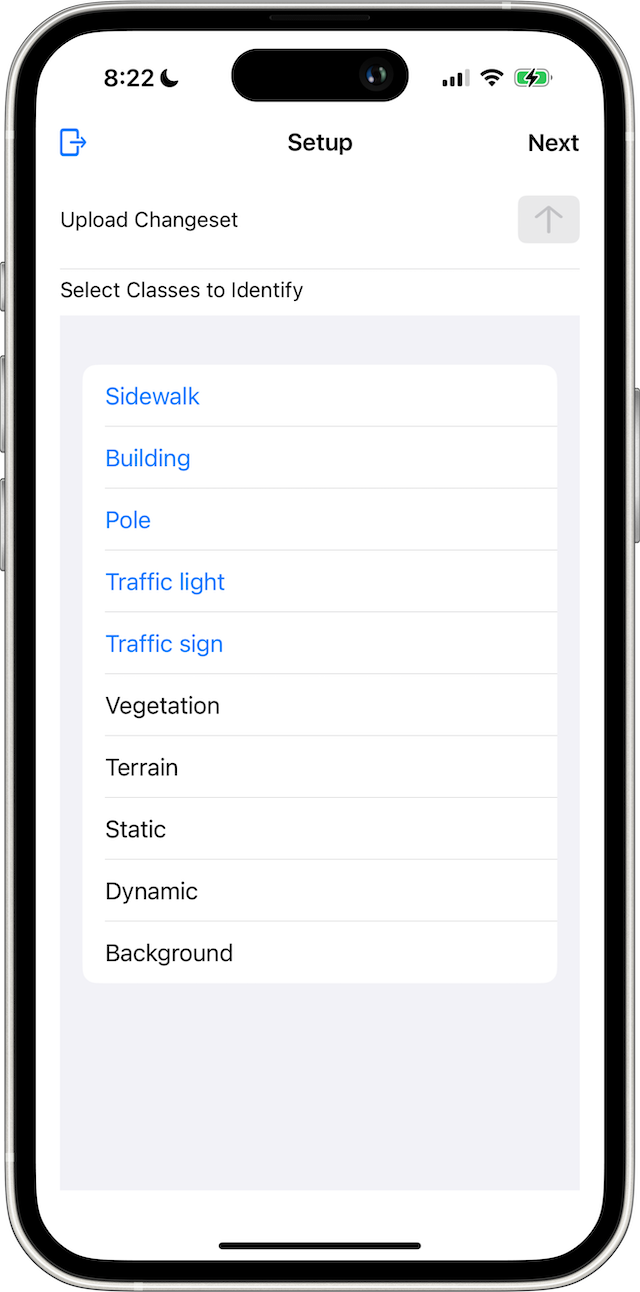}
    
    \caption{}
    \label{fig:login_setup_camera_screenshots_b}
    \end{subfigure}
    \hfill
    \begin{subfigure}[b]{0.3\columnwidth}
    \includegraphics[width=\columnwidth]{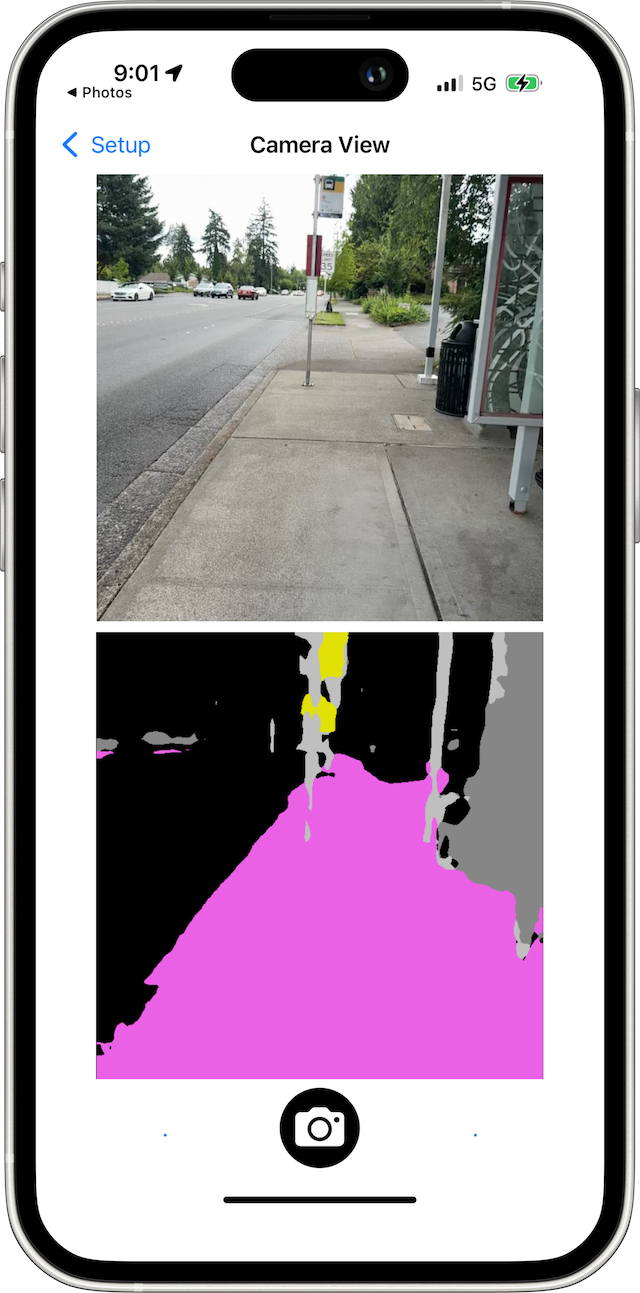}
    \caption{}
    \label{fig:login_setup_camera_screenshots_c}
    \end{subfigure}
  
  %\begin{tabular}{ccc}
  %  \includegraphics[width=0.3\linewidth]{LoginView.png} &
  %  \includegraphics[width=0.3\linewidth]{SetupView.png} &
   % \includegraphics[width=0.3\linewidth]{CameraView_Edited.png} \\
    %\small (a) &
    %\small (b) &
    %\small (c)
  %\end{tabular}
  \caption{(a) Login View: where users registered with the TDEI platform can log into the application (b) Setup View: where users select which feature classes to localize, and also deal with changesets for the TDEI API, (c) Camera View: to display the live camera feed alongside real-time segmentation masks of selected feature classes.}
  \label{fig:login_setup_camera_screenshots}
\end{figure}

To facilitate user interaction with the feature detection pipeline, iOSPointMapper presents users with a Setup View (see Figure \ref{fig:login_setup_camera_screenshots_b}), which serves as the landing screen for configuring a mapping session. This interface displays a list of supported feature classes that the model is capable of detecting and localizing. Users can select one or more classes based on their data collection goals before proceeding. Once the desired configuration is selected, the application moves to the Camera View (see Figure \ref{fig:login_setup_camera_screenshots_c}), where the real-time feature detection process begins. The top portion of the screen displays a live camera feed of the surrounding environment, while the bottom portion overlays segmentation masks corresponding to detected instances of the selected classes. This layout enables users to validate the detections before capturing a frame. When the user is satisfied with the segmentation output, they can initiate capture, which then transitions to the Annotation View for post-capture validation and editing (detailed in Section \ref{sub_methodology_vetting}).

\subsection{Feature Localization and Spatial Mapping} \label{sub_methodology_feature_localization}

\textit{iOSPointMapper} localizes each detected feature instance in geographic co-ordinates by leveraging the depth map from the LiDAR sensor and the pose estimation capabilities available through an ARKit session. ARKit is Apple’s augmented reality framework for iOS devices that, combines camera imagery, IMU, and additional device sensors to estimate the camera’s position and orientation relative to a world co-ordinate system. When the Camera View of \textit{iOSPointMapper} is initialized, it starts an ARKit session, which establishes a world coordinate system and continuously tracks the device relative to this system's origin. The camera's 3D position and orientation details are provided as a \(4\times 4\) matrix called the camera pose. ARKit also provides the camera intrinsics matrix, containing internal calibration parameters such as focal length and principal point, which define how 3D points in camera space project onto the 2D image plane. At the moment of image capture, the application records a synchronized set of sensor inputs: (i) the segmentation mask of detected features from the camera image, (ii) the LiDAR depth map, (iii) the camera intrinsics, (iv) the camera pose, and (v) the device’s GPS coordinates. 

For each detected object, the contour extracted from the segmentation mask is used to compute its pixel centroid. To improve robustness to depth noise, the system estimates depth as the average of LiDAR values within a small circular neighborhood around the centroid rather than relying on a single pixel. Using the camera intrinsics, this pixel location and depth value are back-projected into a 3D point in the camera coordinate system. The camera pose is then applied to transform this point into ARKit’s world coordinate frame, producing a metric 3D position of the feature relative to the device’s environment.

To obtain geographic coordinates, the relative 3D displacement between the device and the feature is projected onto the horizontal plane. This planar displacement is then translated into latitude and longitude offsets from the device’s GPS location. Using the device location as reference, the feature's geographical offsets are then translated into geographic coordinates via spherical projection \citep{Chris_Veness}.

This pipeline ensures orientation-invariant localization, enabling users to hold their device freely while mapping. The geo-location of the features is also done entirely on-device, eliminating the need of transmitting any PII to external servers. 

% \begin{algorithm}[t!]
% \caption{Feature Localization Algorithm}
% \label{alg:localization}
% \KwIn{
%     Segmentation mask $S \in \mathbb{Z}^{w \times h}$;\\
%     Object contour $\Gamma$ (list of contour pixel coordinates);\\
%     Depth map $D \in \mathbb{R}^{w \times h}$;\\
%     Intrinsics $K \in \mathbb{R}^{3 \times 3}$;\\
%     Pose $(R, t)$;\\
%     Device GPS $(\phi_0, \lambda_0)$,\\
%     Radius $r$ (in pixels)
% }
% \KwOut{Feature GPS coordinates $(\phi, \lambda)$}
% \vspace{0.5em}

% Extract object sub-mask $M \subseteq S$\ from contour $\Gamma$\;;
% Compute centroid $p = (u, v) \gets \frac{1}{|M|} \sum_{(i,j)\in M} (i, j)$\;
% Define circular region $\mathcal{C}_r = \{(i,j) \in M \mid \sqrt{(i - u)^2 + (j - v)^2} \leq r \}$\;
% Compute average depth: $d \gets \frac{1}{|\mathcal{C}_r|} \sum_{(i,j) \in \mathcal{C}_r} D(i,j)$\;
% Back-project pixel: $x_{cam} \gets d \cdot K^{-1} [u, v, 1]^T$\;
% Transform to world: $x_{world} \gets R \cdot x_{cam} + t$\;
% Compute delta: $x_\Delta \gets x_{world} - t$\;
% Project to 2D: $(\Delta_\phi, \Delta_\lambda) \gets \text{extract planar components from } x_\Delta$\;
% Compute polar form: $d \gets \sqrt{\Delta_\phi^2 + \Delta_\lambda^2}$, $\theta \gets \arctan2(\Delta_\lambda, \Delta_\phi)$\;
% Compute angular distance: $\delta \gets d / R_e$\;
% Calculate new latitude:
% $\phi \gets \arcsin( \sin \phi_0 \cos \delta + \cos \phi_0 \sin \delta \cos \theta )$\;
% Calculate new longitude:
% $\lambda \gets \lambda_0 + \arctan2( \sin \theta \sin \delta \cos \phi_0, \cos \delta - \sin \phi_0 \sin \phi )$\;

% \Return $(\phi, \lambda)$\;
% \end{algorithm}

\subsubsection{Sidewalk Localization and Geometric Estimation}

For the sidewalk feature class, \textit{iOSPointMapper} supports both geographic localization and the estimation of geometric properties. To focus on the segment the user is currently traversing, the application limits analysis to a fixed Region of Interest (ROI), that is pre-determined by analysing confidence maps of the deployed model. Pixels labeled as sidewalk but falling outside this region are excluded. Within the ROI, the system computes the largest valid trapezoidal region that approximates the portion of the sidewalk directly in front of the user. This region is identified using a top-down bounded traversal, which evaluates horizontal lines from the top to the bottom of the mask to determine the upper and lower bounds of a coherent trapezoid. The centroid of this trapezoid is then used to localize the sidewalk segment, following the same ARKit-based procedure as with other object classes.

The geometric bounds of the trapezoid enable physical measurement of the sidewalk width. The sidewalk width is estimated by projecting the top and bottom edges of the trapezoid into geographic coordinates and averaging their lengths. 

\begin{figure}[t!]
  \centering
    \begin{subfigure}[b]{0.28\columnwidth}
        \includegraphics[width=\columnwidth]{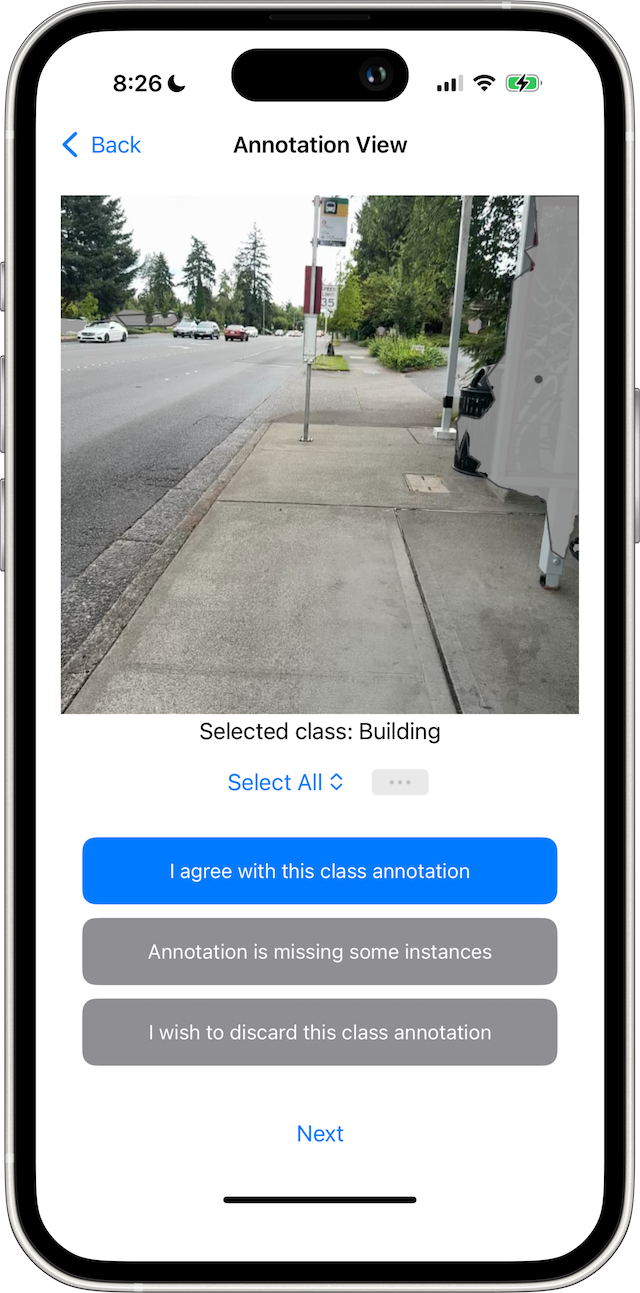}
        \caption{}
        \label{fig:annotation_view_a}
    \end{subfigure}
    \hfill
    \begin{subfigure}[b]{0.28\columnwidth}
        \includegraphics[width=\columnwidth]{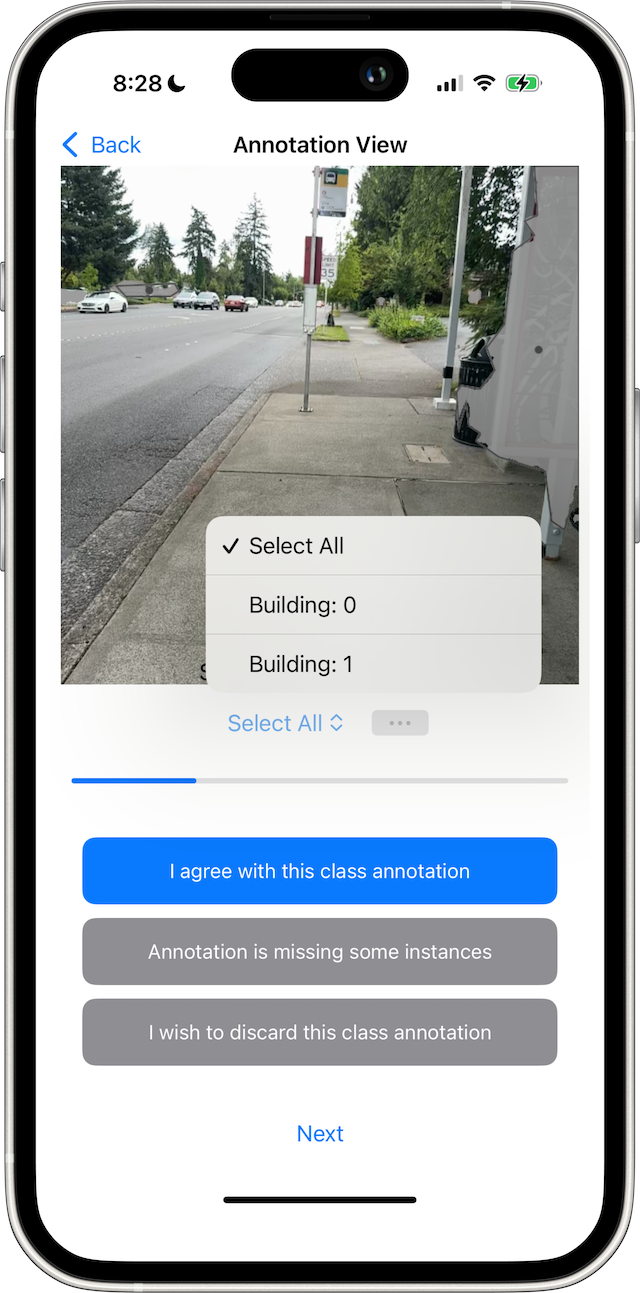}
        \caption{}
        \label{fig:annotation_view_b}
    \end{subfigure}
    \hfill
    \begin{subfigure}[b]{0.28\columnwidth}
        \includegraphics[width=\columnwidth]{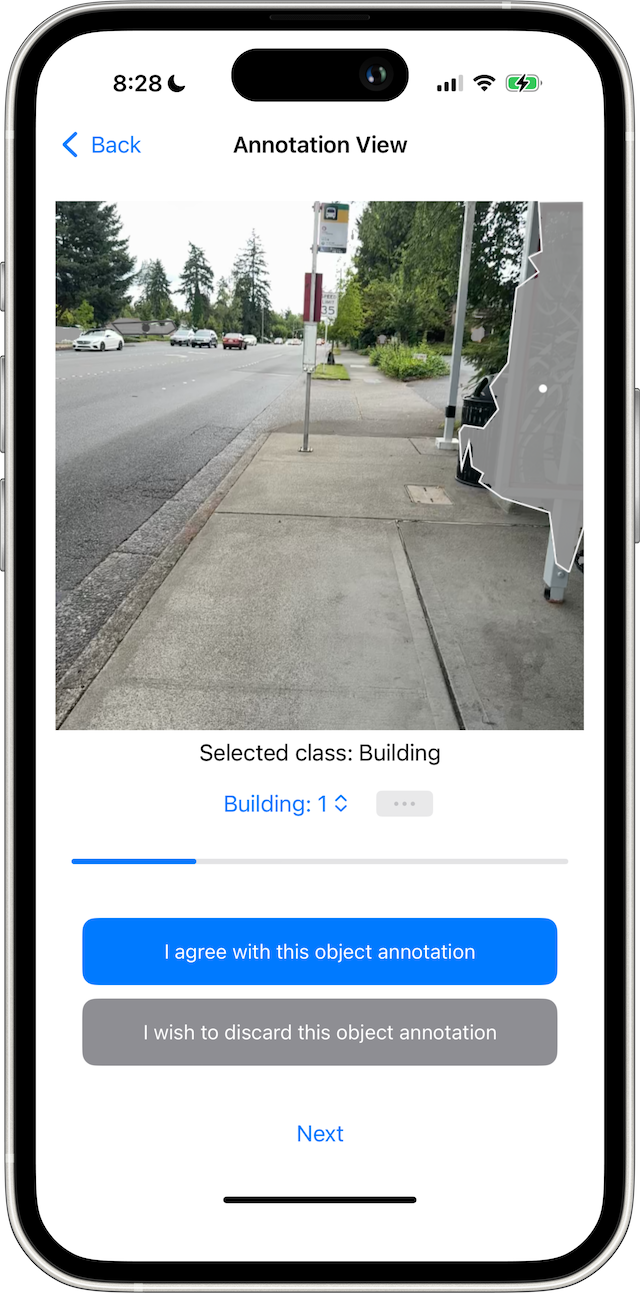}
        \caption{}
        \label{fig:annotation_view_c}
    \end{subfigure}
    \vfill
    \begin{subfigure}[b]{0.49\columnwidth}
        \centering
        \includegraphics[width=0.59\columnwidth]{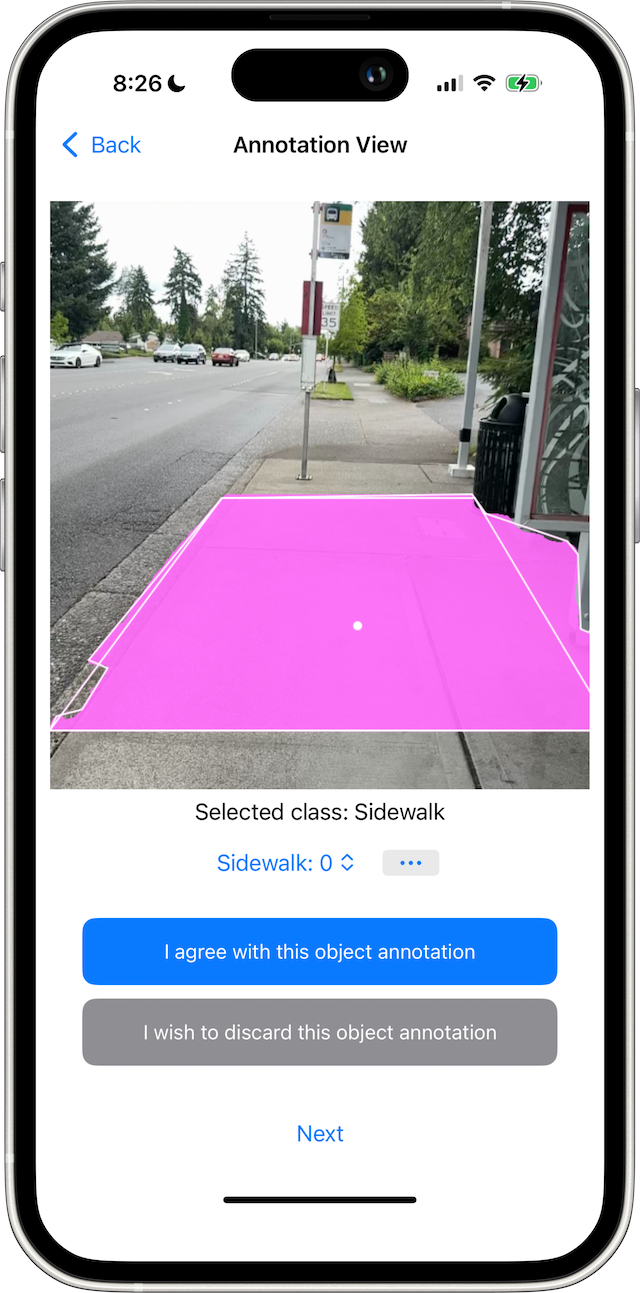}
        \caption{}
        \label{fig:annotation_view_d}
    \end{subfigure}
    \hfill
    \begin{subfigure}[b]{0.49\columnwidth}
        \centering
        \includegraphics[width=0.59\columnwidth]{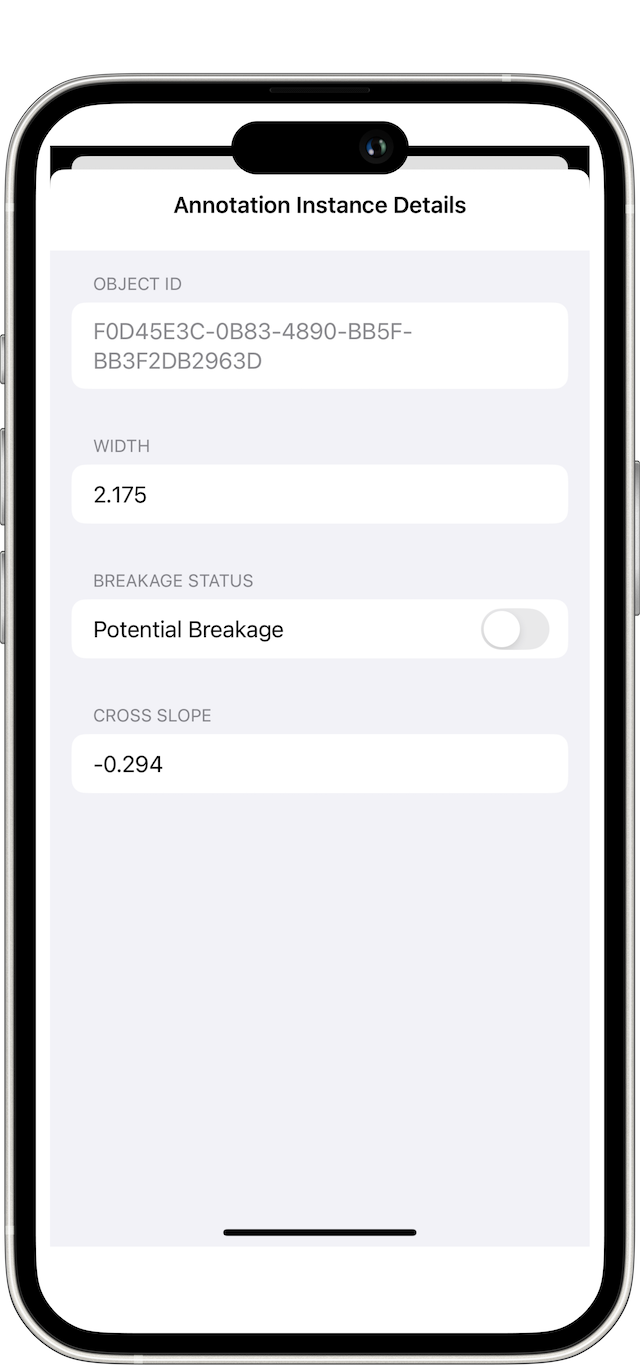}
        \caption{}
        \label{fig:annotation_view_e}
    \end{subfigure}
  \caption{Annotation View: (a) Options to Validate all Instances at once, (b) Selecting Individual Instances for Validation, (c) Options to Validate an Individual Object Instance, (d) Validating Sidewalk Segment, (e) Options to Validation attributes of a Sidewalk Segment}
  \label{fig:annotation_view}
\end{figure}

\subsection{User-Guided Vetting of Detected Features} \label{sub_methodology_vetting}

To ensure a transparent and user-centered mapping experience, \textit{iOSPointMapper} displays an Annotation View (Figure \ref{fig:annotation_view}) after every capture, to allow users to review and validate the mapping results. This view is displayed immediately after the user captures a frame for mapping in the Camera View. By default, user agrees to send all the mapping output. Alternatively, users can choose to reject individual or all instances of a feature class. 

The Annotation Interface shows one screen per detected feature class (e.g., sidewalk, pole, traffic light). As shown in Figure \ref{fig:annotation_view_a},   users are first asked a set of high-level validation options for each class (\textit{Agree} (accept all instances, unless individually rejected), \textit{Discard} (reject all instances), or \textit{Missing} (accept all detected instances unless individually rejected, but flag that some are absent).

For finer control, users can expand a drop-down (Figure \ref{fig:annotation_view_b}) to inspect individual instances, enabling targeted correction. If the user has already selected the user can reject individual instances, while ensuring that the other instances are accepted if user selects Agree for the high-level validation. After completing the review for one class, users proceed to the next via the ``Next'' button, eventually returning to the Camera Interface once all classes have been validated.

This flexible vetting process reinforces a responsible approach to AI deployment, by placing users directly in control of the app's output, to any degree as desired.

\subsection{Integration with TDEI Workspaces} \label{sub_methodology_integration}

To support standardized and centralized storage of sidewalk mapping data, \textit{iOSPointMapper} integrates directly with the TDEI platform. The TDEI platform provides a structured framework for transmitting, validating, and maintaining multimodal transportation datasets. It also provisions a shared API and a dedicated interface called TDEI Workspaces, to manage and interact with the TDEI datasets. Users registered with the TDEI platform can log in through the Login View (Figure \ref{fig:login_setup_camera_screenshots_a}) of \textit{iOSPointMapper}, which authenticates access to a specific workspace in the TDEI platform.

Once a user is authenticated, a new \textit{changeset} is opened for the TDEI workspace. A changeset represents a logically grouped collection of edits by a single contributor. As a user validates features within the app, each accepted object instance is transmitted to the TDEI workspace as a distinct node, containing structured metadata such as node location, changeset ID, and user ID. At any time, the user may close the current changeset—finalizing the associated edits—and initiate a new one through the Setup View (Figure \ref{fig:login_setup_camera_screenshots_b}). The user can later view and edit these changes by accessing the workspace in TDEI Workspaces. 

\subsubsection{Sidewalk Data Transmission}
iOSPointMapper, like other tools in the TDEI ecosystem, use the OpenSidewalks (OSW) data model \citep{OpenSidewalks}. OSW's data model represents sidewalks as line geometries, meaning an ordered sequence of nodes that together describe a continuous walking surface. To align with this convention, iOSPointMapper transmits detected sidewalks not as isolated point observations but as way elements. During the collection, the app creates node elements for sidewalk segment frames that the user traversed. These are enhanced with GPS coordinates and attributes such as detected width. These nodes are then linked in sequence to form a single connected sidewalk way. The sequencing is managed through a changeset, which is simply a bounded editing session that groups all edits recorded by the iOSPointMapper during a single mapping session. When the user finishes mapping a sidewalk and moves on to a different one, a changeset is closed and a new one begins. This approach ensures that the data submitted to the TDEI ecosystem preserves the standardization of the data (through OSW), the geometry of the sidewalk and the temporal continuity of the mapper's session. Future iterations of our system may allow users to segment or merge sidewalks without relying on changeset boundaries, allowing finer editorial control while maintaining OSW compatibility.

A key architectural decision in iOSPointMapper is the execution of the OSW feature extraction pipeline directly on the mobile device. All segmentation, object identification, and geometry construction occur locally, and only the resulting sparse traversal abstraction of the pedestrian infrastructure is transmitted in OSW format. In practical terms, the system discards raw imagery after inference and retains only the geo-referenced sidewalk nodes and associated attributes needed to assemble OSW-compliant ways. This design eliminates the need to upload high-bandwidth image or video streams to a cloud service for computer vision processing, thereby removing a significant transmission bottleneck. It also provides predictable upload requirements in field settings with limited connectivity and reduces latency between data capture and availability for downstream applications such as routing and accessibility analytics. By transmitting only the essential map primitives rather than the underlying sensor data, the app supports both scalable field deployment and strong privacy protections for passers-by and adjacent properties.

% \hfill\break

\section{Experiments} \label{experiments}

\subsection{Hardware Configuration}

All experiments were conducted using an iPhone 16 Pro device, equipped with standard onboard sensors including a LiDAR scanner, a triple-lens pro camera system (with a 48MP main camera, 48MP ultrawide camera, and 12MP 5x telephoto camera), GPS, and IMU. The app leverages these sensors natively via Apple’s ARKit and CoreLocation APIs to capture RGB-D frames, camera pose, and geolocation data during mapping.

\subsection{Feature Detection Experiments} \label{sub_experiments_feature_recognition}

The evaluation of the feature detection pipeline focuses on the semantic segmentation component, as it forms the backbone of the pipeline. We benchmarked three BiSeNetv2 variants, each trained with a different strategy. The first model, \textit{BiSeNetv2-City}, was trained solely on the Cityscapes dataset, which focuses on vehicular urban environments. The second variant, \textit{BiSeNetv2-PED}, following the methodology proposed in \citep{zhang2023oasis}, was pretrained on the COCO-Stuff dataset and then fine-tuned on the custom pedestrian-view dataset described in Section \ref{sub_methodology_feature_recognition}. These two models serve as baseline models in this experiment. The third and final model, \textit{BiSeNetv2-iOS}, was pretrained on the curated subset of Mapillary Vistas as mentioned in Section \ref{sub_methodology_feature_recognition}, and subsequently fine-tuned on the same custom pedestrian dataset. This final variant is the one deployed in the application. All models were evaluated on a held-out validation split comprising 20\% of the annotated dataset.

\subsection{Mapping Experiments}
\label{sub_experiments_feature_localization}

We conducted field tests in three cities of Washington State—Seattle, Bellevue, and Redmond—to evaluate the full mapping pipeline under diverse urban conditions. For each city, a sidewalk stretch of approximately 3 kilometers was selected. A single annotator used the \textit{iOSPointMapper} application—equipped with the \textit{BiSeNetv2-iOS} model—to perform continuous, ground-level mapping along the route. For small object classes: pole, traffic sign and traffic pole, The collected data was uploaded to a designated TDEI workspace and later analyzed to assess localization accuracy, sidewalk measurement accuracy, and mapping efficiency as detailed in Section~\ref{results}.

\section{Results} \label{results}

\subsection{Feature Detection Performance}
\label{sub_results_feature_recognition}

As described in Section \ref{sub_experiments_feature_recognition}, the performance of the feature detection pipeline was evaluated by benchmarking three BiSeNetv2 variants trained under different data strategies. In this section, we present the experimental results, focusing on the model’s ability to segment the pedestrian-relevant features with high precision and reliability under varied real-world conditions.

Table \ref{table_segmentation_metrics} summarizes the segmentation results across the five feature classes, evaluated on the held-out validation split of our annotated pedestrian-view dataset. We report class-wise  Intersection over Union (IoU), Precision, and Recall — metrics that reflect both per-class accuracy and the robustness required for sidewalk accessibility mapping. 

\begin{table}[ht]
\centering
\resizebox{\textwidth}{!}{%
\begin{tabular}{|l|ccc|ccc|ccc|ccc|ccc|}
\toprule
% \textbf{Model} 
&
\multicolumn{3}{c|}{\textbf{Sidewalk}} &
\multicolumn{3}{c|}{\textbf{Building}} &
\multicolumn{3}{c|}{\textbf{Traffic Sign}} &
\multicolumn{3}{c|}{\textbf{Traffic Light}} &
\multicolumn{3}{c|}{\textbf{Pole}}\\
\textbf{Model} & IoU & Precision & Recall & IoU & Precision & Recall & IoU & Precision & Recall & IoU & Precision & Recall & IoU & Precision & Recall\\
\midrule
BiSeNetv2-City & 0.0881 & 0.5132 & 0.0962 & 0.5098 & 0.7057 & 0.6475 & 0.1684 & 0.6187 & 0.1879 & 0.3023 & 0.7383 & 0.3386 & 0.0749 & 0.8134 & 0.0762\\
BiSeNetv2-PED & 0.7159 & 0.7367 & 0.9621 & \textbf{0.6149} & \textbf{0.6508} & 0.9176 & 0.3829 & 0.4902 & 0.6364 & 0.2793 & \textbf{0.7735} & 0.3042 & 0.3951 & 0.4742 & 0.7031\\
BiSeNetv2-iOS & \textbf{0.7782} & \textbf{0.7968} & \textbf{0.9708} & 0.6067 & 0.6342 & \textbf{0.9334} & \textbf{0.5182} & \textbf{0.6638} & \textbf{0.7026} & \textbf{0.4636} & 0.7087 & \textbf{0.5727} & \textbf{0.4379} & \textbf{0.5331} & \textbf{0.7103}\\
\bottomrule
\end{tabular}
}
\caption{Segmentation Performance by Object Class and Model Variant}
\label{table_segmentation_metrics}
\end{table}

For the sidewalk class, BiSeNetv2-iOS—trained with pedestrian-centric pretraining and fine-tuning—shows substantial gains over both baselines. Visual examples in Figure \ref{fig:segmentation_results_images} show how BiSeNetv2-City frequently misclassifies sidewalk surfaces as road, while both BiSeNetv2-PED and BiSeNetv2-iOS correctly segment the ground plane as sidewalk, as expected in our deployment context. These improvements directly support more accurate downstream geometric estimation.

For other classes—buildings, traffic signs, traffic lights, and poles—model performance improves progressively with increased training alignment to the pedestrian viewpoint. For building, BiSeNetv2-PED performs slightly better, most likely due to the much larger size of its pretraining dataset. Traffic light and pole detection remain the most challenging due to occlusion and clutter, indicating the need for more context-aware refinement in future work. Overall, BiSeNetv2-iOS demonstrates strong generalization across all POI types, validating its use in iOSPointMapper for robust, mobile-friendly feature detection.

Together, these results confirm that pedestrian-centric pretraining and annotation are critical for feature detection in sidewalk environments. BiSeNetv2-iOS outperforms both baseline models across a majority of the metrics and object classes, making it the most suitable candidate for deployment with iOSPointMapper.

\definecolor{magenta}{RGB}{244, 35, 232}
\definecolor{darkgrey}{RGB}{70, 70, 70}
\definecolor{lightgrey}{RGB}{153, 153, 153}
\definecolor{lightyellow}{RGB}{220, 220, 0}
\definecolor{darkyellow}{RGB}{250, 170, 30}
\begin{figure}
  \centering
  \begin{tabular}{c}
    \includegraphics[width=0.95\linewidth]{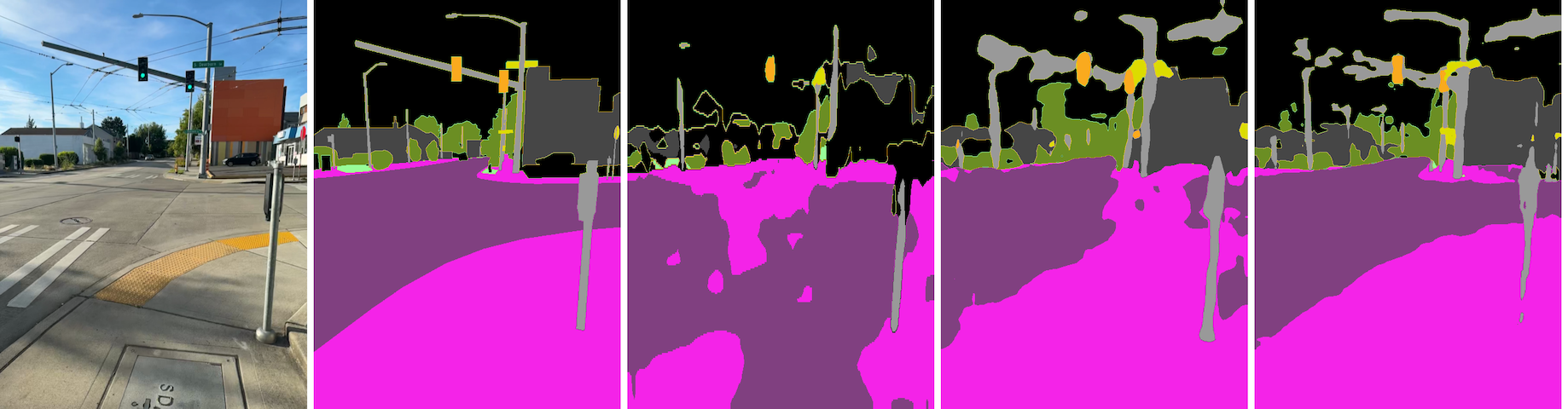} \\
    \includegraphics[width=0.95\linewidth]{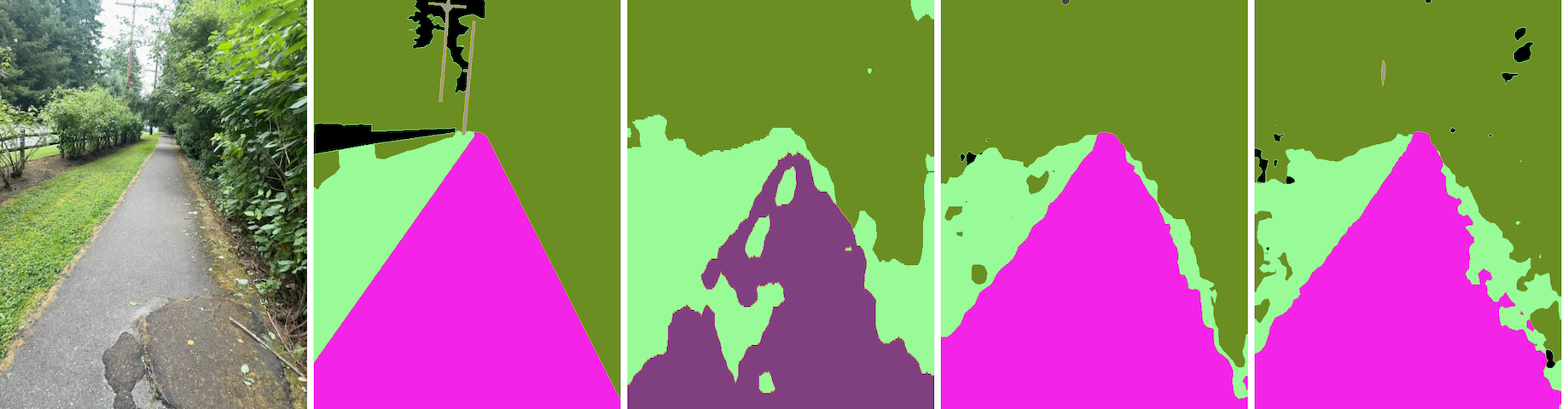} \\
    \includegraphics[width=0.95\linewidth]{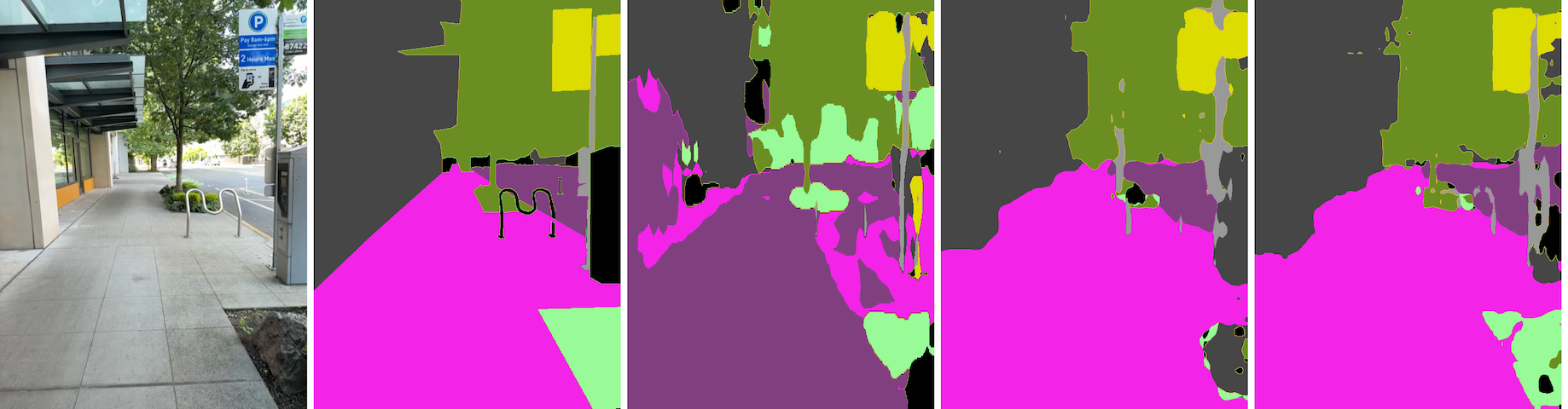} \\
    \includegraphics[width=0.95\linewidth]{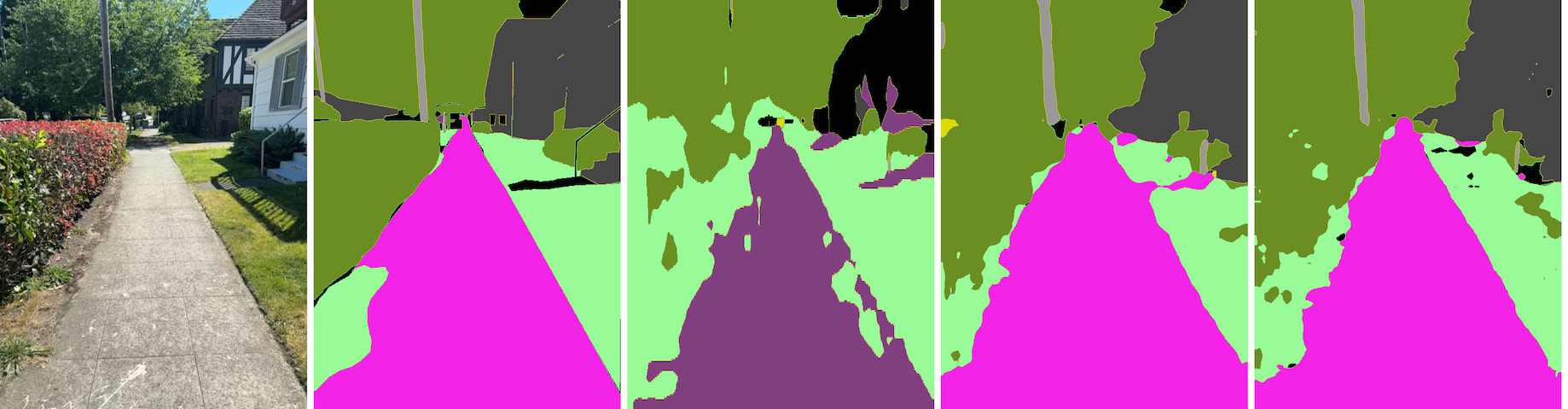} \\
  \end{tabular}
  \caption{Segmentation Results of various Models on Sidewalk-Imagery captured by iPhone. \textbf{Column 1}: Input Image, \textbf{Column 2}: Ground-truth Segmentation Masks, \textbf{Column 3}: Segmentation Output produced by BiSeNetv2-City, \textbf{Column 4}: Segmentation Output produced by BiSeNetv2-PED, \textbf{Column 5}: Segmentation Output produced by BiSeNetv2-iOS. \\
  In the segmentation masks, sidewalks are represented in \colorbox{magenta}{\textcolor{white}{magenta}}, buildings in \colorbox{darkgrey}{\textcolor{white}{dark grey}}, traffic signs in \colorbox{lightyellow}{light yellow}, traffic lights in \colorbox{darkyellow}{\textcolor{white}{dark yellow}}, and finally poles in \colorbox{lightgrey}{light grey}.}
  \label{fig:segmentation_results_images}
\end{figure}

% \FloatBarrier
\subsection{Mapping Performance}

This section details the results of mapping experiments as described in Section \ref{sub_experiments_feature_localization}. These evaluations measured the system’s ability to localize identified features and estimate sidewalk geometry with accuracy suitable for accessibility-focused use cases.

Quantitative results are presented in Table~\ref{table_localization_erros} and Table~\ref{table_measurement_errors}. For each object class, we report localization errors in terms of mean, standard deviation, and root mean square error (RMSE), with sidewalk width estimation evaluated as a separate metric. The reported values account for device-level GPS uncertainty to isolate the effect of the visual pipeline.

\begin{table}[ht]
\centering
\resizebox{\textwidth}{!}{%
\begin{tabular}{|l|ccc|ccc|ccc|ccc|}
\toprule
% \textbf{Model} 
&
\multicolumn{3}{c|}{\textbf{Building}} &
\multicolumn{3}{c|}{\textbf{Traffic Sign}} &
\multicolumn{3}{c|}{\textbf{Traffic Light}} &
\multicolumn{3}{c|}{\textbf{Pole}}\\
\textbf{City} & Mean & Std. Dev. & RMSE & Mean & Std. Dev. & RMSE & Mean & Std. Dev. & RMSE & Mean & Std. Dev. & RMSE \\
\midrule
Seattle &
0.8942 & 0.7283 & 1.1275 & 0.9037 & 0.8589 & 1.2135 & 2.7573 & 3.9025 & 4.4482 & 1.2568 & 1.1989 & 1.7258
\\
Bellevue & 0.7124 & 0.5160 & 0.8606 & 0.8546 & 0.8469 & 1.1781 & 2.5253 & 3.2117 & 3.9011 & 1.2729 & 1.8172 & 2.1769
\\
Redmond &
0.9376 & 0.6915 & 1.1420 & 0.9533 & 0.8656 & 1.2632 & 2.5803 & 3.1802 & 3.9149 & 1.2290 & 1.244 & 1.7308
\\
\bottomrule
\end{tabular}
}
\caption{Localization Errors by Object Class and City (in m)}
\label{table_localization_erros}
\end{table}

\begin{table}[h!]
\centering
\resizebox{0.5\textwidth}{!}{%
\begin{tabular}{|l|ccc|ccc|}
\toprule
% \textbf{Model} 
&
\multicolumn{3}{c|}{\textbf{Location}} &
\multicolumn{3}{c|}{\textbf{Width}} \\
\textbf{Model} & Mean & Std. Dev. & RMSE & Mean & Std. Dev. & RMSE \\
\midrule
Seattle 
& 0.5583 & 0.5152 & 0.7579 & 0.9015 & 0.9513 & 1.3070
\\
Bellevue & 0.5670 & 0.4666 & 0.7327 & 0.8417 & 0.8373 & 1.1840
\\
Redmond & 0.5972 & 0.4748 & 0.7614 & 0.8920 & 0.8440 & 1.2249
\\
\bottomrule
\end{tabular}
}
\caption{Measurement Errors of Sidewalks by City (in m)}
\label{table_measurement_errors}
\end{table}

We observe that large, grounded features such as buildings and sidewalks are consistently mapped with low localization error. In contrast, smaller or more distant objects—particularly traffic lights and poles—exhibit higher variance, largely due to limitations in both the segmentation model and LiDAR-based depth estimation at distances beyond 5 meters \citep{TEO2023100169}. Occlusion and clutter contribute further to these difficulties.

Sidewalk width estimation remains within acceptable bounds for pedestrian infrastructure mapping, though errors are amplified in scenes with complex ground geometry or reduced depth visibility. Notably, when isolating only objects within 5 meters of the user, both object localization and sidewalk measurement accuracy improve markedly—often below 0.5 meters RMSE.

These results underscore both the promise and the current limitations of on-device mapping in uncontrolled urban environments. Improving the performance of distant or occluded object localization remains a key challenge. Future enhancements will explore post-processing techniques such as photogrammetry-based reconstruction for better depth assessment.

\FloatBarrier
\subsection{Integration with TDEI Workspaces}

\begin{figure}[h!]
  \centering
  \begin{tabular}{cc}
    \includegraphics[height=0.6\linewidth]{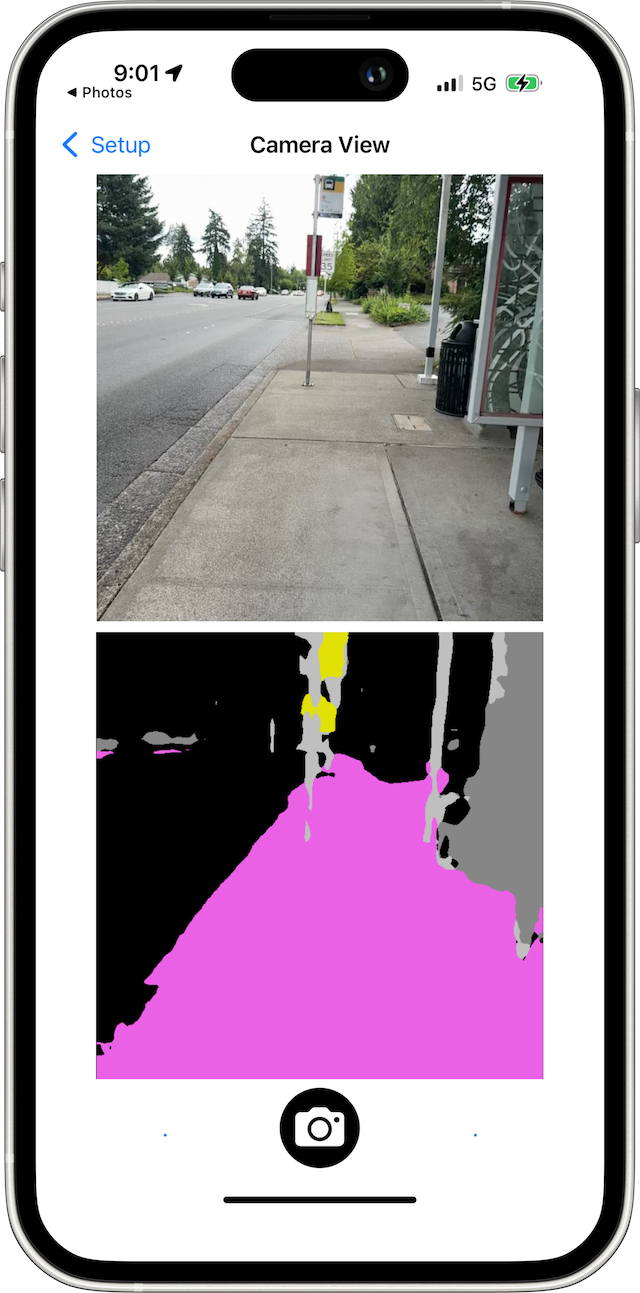} &
    \includegraphics[height=0.4\linewidth]{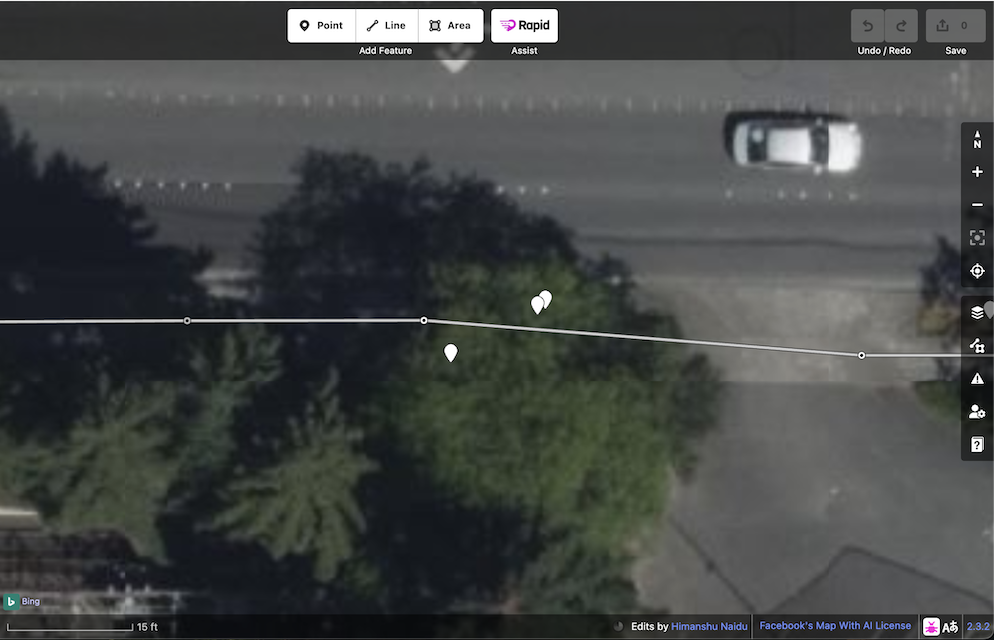} \\
    \small (a) &
    \small (b) \\
    \includegraphics[height=0.6\linewidth]{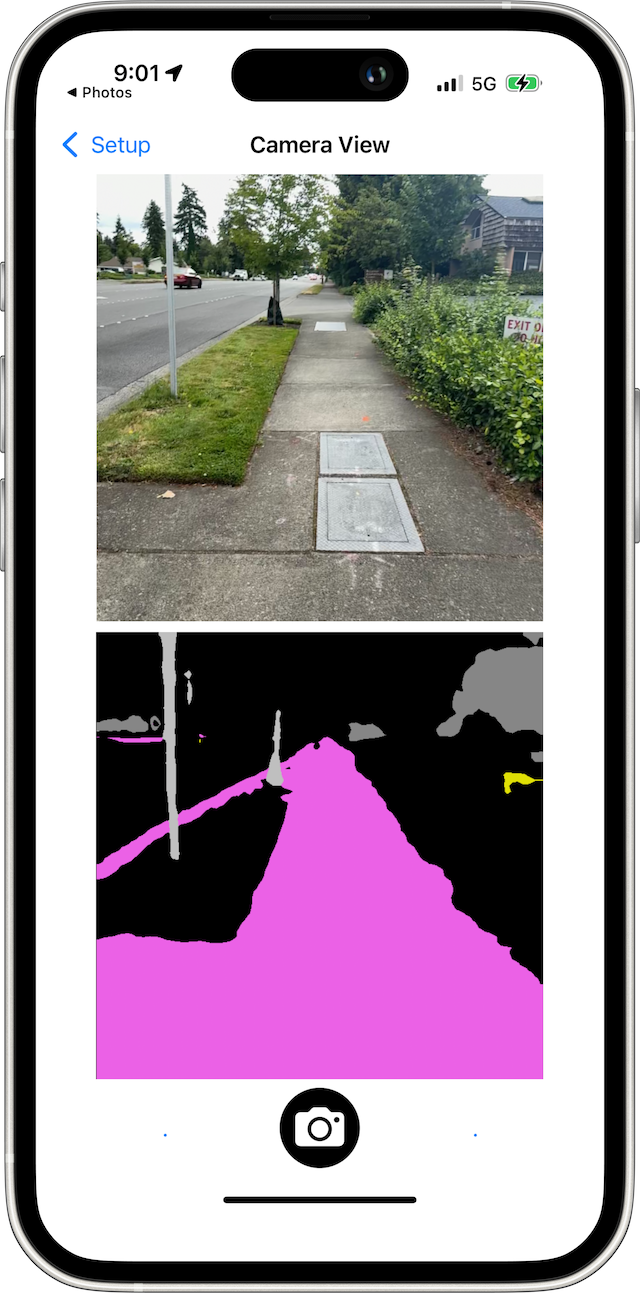} &
    \includegraphics[height=0.4\linewidth]{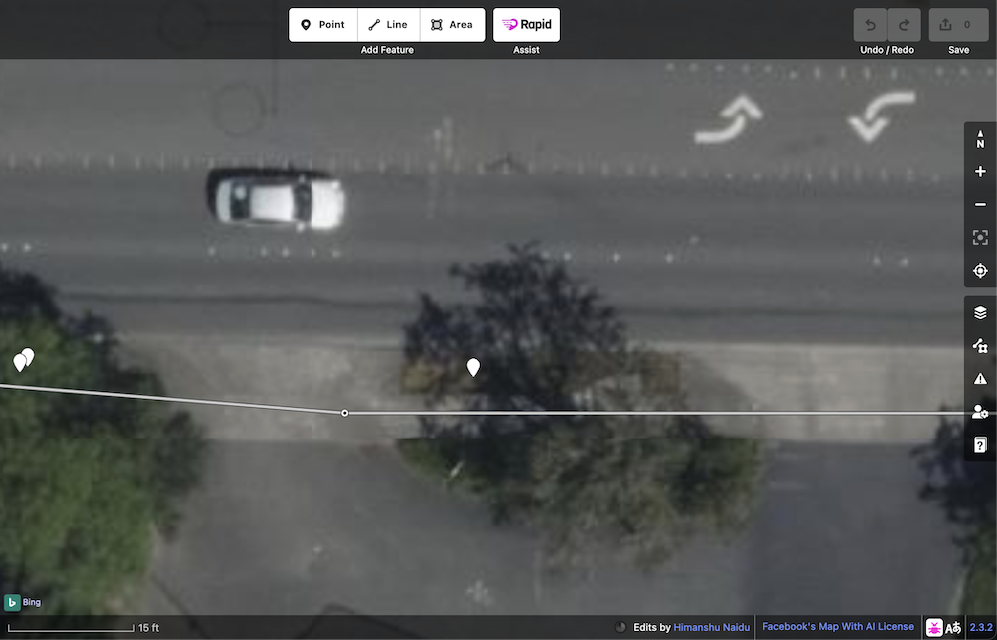} \\
    \small (c) &
    \small (d)
  \end{tabular}
  \caption{Integration with TDEI Workspaces. (a, b) \textit{iOSPointMapper} view and corresponding TDEI workspace of a sidewalk scene with a bus station, traffic light, and two poles.}
  \label{fig:mapping_results_tdei}
\end{figure}

In this section, we visualize the transmitted features within a designated TDEI workspace. Figure~\ref{fig:mapping_results_tdei} shows screenshots from the \textit{iOSPointMapper} app alongside corresponding renderings in the TDEI Workspaces interface.

In Figures \ref{fig:mapping_results_tdei}(a),(b), the user captures a sidewalk containing a bus stop, a pole-mounted traffic sign, and adjacent poles. These features are recognized, localized, and transmitted as individual geotagged nodes. The corresponding sidewalk segment is also tagged as a node. The TDEI Workspaces interface reflects these detections as labeled point features.

Figures \ref{fig:mapping_results_tdei}(c),(d), show a continuation of the same sidewalk stretch. A new pole is captured and accurately mapped to the workspace. The sidewalk itself is recorded as a continuous way-type element, which is visualized as a connected line geometry in the workspace. This supports intuitive inspection of both discrete objects and structural elements like walkable paths.
These examples demonstrate successful end-to-end operation of the mapping pipeline, from on-device recognition to integration with a centralized, standards-compliant data repository.

\FloatBarrier
\section{Discussion} \label{discussion}

\subsection{Implications for Inclusive Sidewalk Navigation}
\textit{iOSPointMapper}, as part of the broader TDEI ecosystem, represents a significant step forward in addressing long-standing challenges in sidewalk infrastructure assessment. By combining real-time, on-device computer vision with mobile sensing, the application offers a compelling response to the data gaps that have historically hindered equitable pedestrian infrastructure development. Crucially, the inclusion of user validation directly on the device fosters trust and ensures higher-quality submissions without introducing PII or dependency on cloud-based processing.

This model allows for a broader set of contributors—including municipal staff, advocacy organizations, and community members—to engage in data collection efforts using off-the-shelf mobile devices. The result is a more participatory mapping process that supports equity-focused infrastructure investment and enables more precise accessibility assessments. 

\subsection{Future Directions}

To improve the robustness and utility of the system, several technical enhancements have been planned. These directions span technical refinements in segmentation and localization, integration with external datasets, and alignment with civic data governance needs.

\subsubsection{Feature Detection Improvements}

Refinement of the on-device segmentation model remains a priority to ensure consistent performance under varied environmental conditions, such as occlusion, poor lighting, cluttered urban environments, or areas with atypical sidewalk geometries. Enhancing the model’s generalization across regions will reduce the dependence on post-processing for correction. 

Beyond segmentation accuracy, future iterations of the system aim to expand the supported feature set. Notably, feature classes such as curb ramps are critical to accessible pedestrian infrastructure and it would be prudent to support these as new semantic classes. 

\subsubsection{Geographic Localization Enhancement}

Ongoing work will also focus on improvements in geographic accuracy, especially for small static objects that the application has struggled to accurately localize. Techniques such as photogrammetry may be employed to improve localization of distant objects to compensate for the limitations of LiDAR scanners. Furthermore, techniques such as dead reckoning may be introduced to maintain localization reliability, particularly in dense urban environments with variable GPS quality. 

The additional sensor capabilities of the iOS devices can be leveraged to improve the sidewalk scene understanding. Future iterations will aim to support estimation of sidewalk attributes such as slope and cross-slope, as well as the identification of surface integrity issues like cracking or heaving. These attributes are critical for ADA accessibility compliance.

\subsubsection{Integration with External Datasets}

Another key future direction involves supporting integration with existing municipal asset inventories, such as previously cataloged poles, signage, or curb infrastructure. Allowing agencies to overlay their pre-existing Geographic Information System (GIS) datasets into the iOSPointMapper interface could enable in-field validation, help flag outdated information, and reduce redundant data collection. Such a task can be made possible in a standardized manner, within the TDEI ecosystem.

% \hfill
% \break 

Looking forward, the integration of iOSPointMapper data with broader open ecosystems—such as OpenStreetMap or regional GIS platforms—could enhance interoperability and civic engagement. Seamless integration with the TDEI platform ensures that collected data can be used across various planning and compliance workflows, strengthening data interoperability and long-term utility. As the datasets that the app contributes to, grow, and standardization improves, this work will help towards a more open, transparent, and equitable transportation data commons—one that supports not just better sidewalks, but more inclusive cities overall.

\section{Conclusion} \label{conclusion}

In this paper, we presented \textit{iOSPointMapper}, a mobile application designed to support scalable, privacy-conscious, and user-centered sidewalk data collection on the ground. Through real-time on-device processing, human-in-the-loop validation, and seamless integration with the TDEI platform, the system bridges critical data gaps while respecting user agency and data privacy. As development continues, expanding its sidewalk-scene understanding capabilities and interoperability with open data ecosystems will position iOSPointMapper as a key enabler of inclusive, data-driven mobility solutions.

\section{Acknowledgments}

We thank Kohei Matsushima, Sai Sunku, and Seanna Qin for their contributions in the development of an initial prototype of the application. We also recognize Ihor Olkhovatyi, Mariana Piz, and Dmytro Stepanchuk for their work on the TDEI API integration and preliminary feature detection exploration. 

This work was supported by the U.S. Department of Transportation (USDOT) Intelligent Transportation Systems for Underserved Communities (ITS4US) Deployment Program, administered by the Joint Program Office (JPO), under Cooperative Agreement Number 693JJ32250014.

\bibliographystyle{trb}
\bibliography{references}
\end{document}